\documentclass[11pt]{article}
\usepackage{acl2014}
\usepackage{times}
\usepackage{url}
\usepackage{latexsym}
\usepackage{graphicx}
\usepackage{url}
\usepackage{caption}
\usepackage{subcaption}
\usepackage{pbox}

\title{Image Representations and New Domains in Neural Image Captioning}

\author{
Jack Hessel\\
Computer Science Dept\\
Cornell University\\
\texttt{jhessel@cs.cornell.edu} \\
\And
Nicolas Savva\\
Computer Science Dept\\
Cornell University\\
\texttt{nss45@cornell.edu} \\
\And
Michael J. Wilber\\
Cornell Tech\\
Cornell University\\
\texttt{mwilber@mjwilber.org}\\
}

\date{}

\begin{document}
\maketitle
\begin{abstract}
  We examine the possibility that recent promising results in
  automatic caption generation are due primarily to language
  models. By varying image representation quality produced by a
  convolutional neural network, we find that a state-of-the-art neural
  captioning algorithm is able to produce quality captions even when
  provided with surprisingly poor image representations. We replicate
  this result in a new, fine-grained, transfer learned captioning
  domain, consisting of 66K recipe image/title pairs. We also provide
  some experiments regarding the appropriateness of datasets for
  automatic captioning, and find that having multiple captions per
  image is beneficial, but not an absolute requirement.
\end{abstract}

\section{Introduction}
Describing the content of an image is an easy task for humans, but,
until recently, had been difficult or impossible for computers. Recent
work in computer vision has addressed this task of automatically
generating the caption of an input image with promising results
\cite{farhadi2010every,kulkarni2013babytalk,ordonez2011im2text,karpathy2014deep,mao2014explain,vinyals2014show,kiros2014unifying,donahue2014long,fang2014captions}. Several
state-of-the-art approaches couple a pre-trained deep convolutional
neural network (CNN) for image representation with a recurrent neural
network (RNN) to generate captions that describe image content.

We consider the possibility that the generation of these captions,
however, is not heavily reliant upon the image representation
input. For instance, if one was to train a RNN directly on image
captions, one could learn a fair amount about the general language of
image captions. Sutskever et al. \shortcite{sutskever2011generating}
demonstrate that RNNs are capable of producing diverse and
surprisingly readable sentences, given a short starting sequence of
seed words. Furthermore, non-neural memoization techniques like those
proposed by Wood et al. \shortcite{wood2009stochastic} and Gasthaus et
al. \shortcite{gasthaus2010lossless} are capable of producing very
convincing language models for particular domains.

While it is clear that existing algorithms do discriminate based on
image inputs, it is still unclear if the apparently highly specific
generated captions are primarily a result of language modeling rather
than image modeling. If it could be determined that either image
modeling or language modeling is acting as the bottleneck in this
multimodal setting, research efforts could be directed
appropriately.

To examine the relative multimodal modeling capacities of existing
neural captioning algorithms, we execute a series of experiments where
we vary image representation quality produced from a fixed CNN, and
examine how the output captions are affected.

For two existing datasets and a new domain we analyze here, our
results suggest that caption quality does not scale well with
increased classification accuracy of a fixed CNN. In fact, as the
testing/validation accuracy of a CNN with fixed architecture
increases, all seven caption evaluation metrics we consider appear to
saturate at surprisingly low classification accuracies. While this
does not prove that better image modeling algorithms could not produce
better captions, it appears that many apparently fine-grained aspects
of generated natural language are the result of surprisingly coarse
grained visual distinctions.

For a fixed vision model, our results indicate that there is likely
little room for caption improvement via gathering more training images
alone. We further postulate that progress could be made most quickly
through the development of language modeling techniques that take
better advantage of existing image representations. In particular,
coupling our results with independent but consistent observations made
by Karpathy and Li \shortcite{karpathy2014deep} and Vinyals et
al. \shortcite{vinyals2014show} regarding model modifications that
lead to overfitting, it's very likely that overfitting language models
to image features is still a big problem for many caption generation
algorithms. Our analysis highlights what we believe to be an important
question for these types of algorithms going forward: if better image
representations contain useful, fine-grained information, is it
possible to take advantage of that information without overfitting?

To supplement our analysis of image representations, we consider a new
caption generating task: generating recipe titles based on images of
food. The motivation for this new task results from the intuition that
image representations might matter more in visually fine-grained
domains, where algorithms must be able to discriminate between minute
changes in the input images. We collect a dataset consisting of images
of food coupled with recipe titles (e.g. ``thai chicken curry'') from
{\tt Yummly.com} for this purpose. When compared to captioning the
coarse-grained ImageNet domain, the specificity of our food dataset
calls for more subtle visual discrimination.

Instead of learning a food image representing CNN from scratch to
derive representations, we apply transfer learning on a dataset of
101K food images. Using this approach, we significantly surpass
current state-of-the-art performance for a classification task on this
dataset, despite using a somewhat outdated deep architecture. We
further demonstrate that this transfer learning process does indeed
improve food captioning, though we observe a similar ``flattening'' of
all linguistic evaluation metrics, after a point.

\section{Related Work}

\subsection{Automatic Captioning}
The model we choose to analyze in detail is the ``Neural Image
Captioning'' (NIC) model detailed by Vinyals et
al. \shortcite{vinyals2014show}, though we believe the experiments we
address here are relevant to researchers working on distinct but
related models. In a similar fashion to Donahue et
al. \shortcite{donahue2014long} and Karpathy and Li
\shortcite{karpathy2014deep}, NIC feeds a pre-classification
representation of images produced by an architecture like GoogLeNet
\cite{szegedy2014going} or AlexNet \cite{alexnet} to a LSTM recurrent
neural network \cite{hochreiter1997long} for language generation. The
RNN weights are usually trained on datasets consisting of pairs of
images and several corresponding human-generated annotations, such as
Flickr8k \cite{hodosh2013framing}, Flickr30k \cite{young2014image}, or
Microsoft COCO \cite{lin2014microsoft}. The CNN is often pre-trained
on a very large set of images such as ImageNet \cite{deng2009imagenet}
and held fixed while the RNN is trained. For many existing captioning
datasets, ImageNet is a convenient starting point, presumably because
images in most modern captioning datasets are of similar objects.

More complicated caption generation models have also demonstrated
success on several datasets. To the knowledge of the authors, Fang et
al. \shortcite{fang2014captions} hold the current best result (in
terms of BLEU-4) on the MSCOCO official captioning test set, though
Vinyals et al. \shortcite{vinyals2014show} reportedly outperform Fang
et al. on 2/5 evaluation metrics detailed on the MSCOCO captioning
leaderboard.\footnote{\texttt{mscoco.org/dataset/}} Their pipeline
involves training a language model directly on captions and a
discretized image representation consisting of a likely set of objects
in that image. Switching from a finetuned AlexNet \cite{alexnet} to a
finetuned VGG-net \cite{simonyan2014very} improved BLEU-4 by 2.4
points, and METEOR by 1.4 points. Because their image representations
were discrete, it's possible that their language models were less prone
to overfitting. It's not immediately obvious that a similar
improvement would occur for language models that operate on extracted
vector representations of images like NIC, however.

In contrast to the previous approaches that provide their RNNs with a
representation of an image only at the first timestep, Mao et
al. \shortcite{mao2014explain} propose an extension of a single-layer
RNN, dubbed the ``multimodal RNN,'' that feeds a representation of an
image to the RNN at \emph{every} word generation step. Finally, Kiros
et al. \shortcite{kiros2014unifying} propose a model that first uses a
CNN and an RNN to embed an image and its corresponding caption in the
same semantic space, and then feeds vectors from this space into a
``language generating structure content neural language model'', an
extension of a multiplicative RNN that ``disentangles the structure of
a sentence to its content.''

Among models that directly input extracted features to a generating
RNN, it is clear that image representations can be
mishandled. Specifically, several authors note that passing image
representations to the RNN at \emph{every} timestep empirically leads
to worse performance. While Karpathy and Li
\shortcite{karpathy2014deep} do not offer speculation as to why this
is the case, Vinyals et al. \shortcite{vinyals2014show} briefly
mention that this operation leads to over-fitting. These independent
observations demonstrate that it is easy to overfit to image features.

\subsection{Caption Evaluation Metrics}

To evaluate captions, we use BLEU-\{1,2,3,4\} \cite{papineni2002bleu}
METEOR \cite{denkowski:lavie:meteor-wmt:2014} and CIDEr/CIDEr-D
\cite{vedantam2014cider}. BLEU-n is a precision measure over n-grams,
whereas METEOR is a more sophisticated metric that involves the
computation of an alignment between candidate and reference captions;
both were originally conceived in the context of machine
translation. CIDEr/CIDEr-D was created to evaluate captions of images
and focuses on consensus, particularly in cases where there are
multiple reference captions.

\subsection{Recipe Title Prediction Tasks}
\begin{figure}
  \centering \includegraphics[scale=.32]{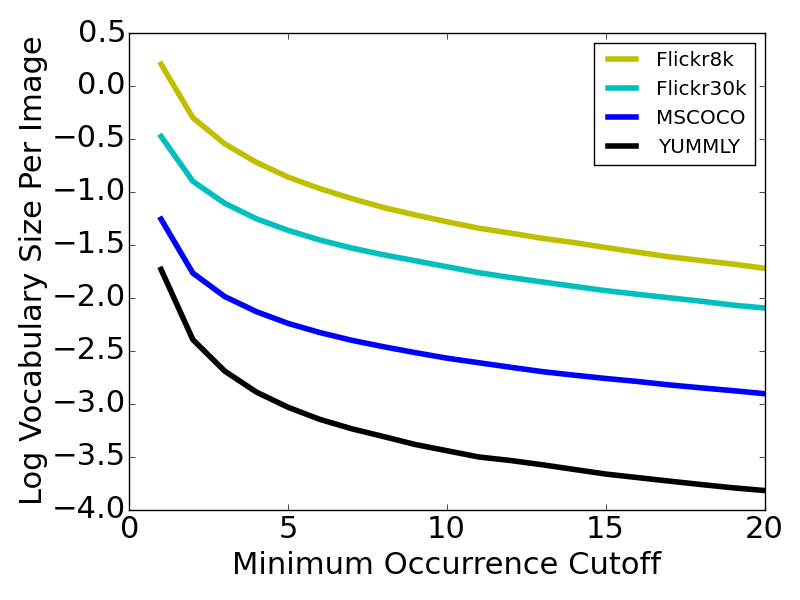}
  \caption{Word cutoff versus log-scale vocab size per
    image. This metric captures both dataset size and
    vocabulary size and shows that Yummly has the smallest 
    vocabulary by a margin.}
    \label{fig:vocabsize}
\end{figure}

\begin{figure*}[t]
\centering
\includegraphics[scale=.8]{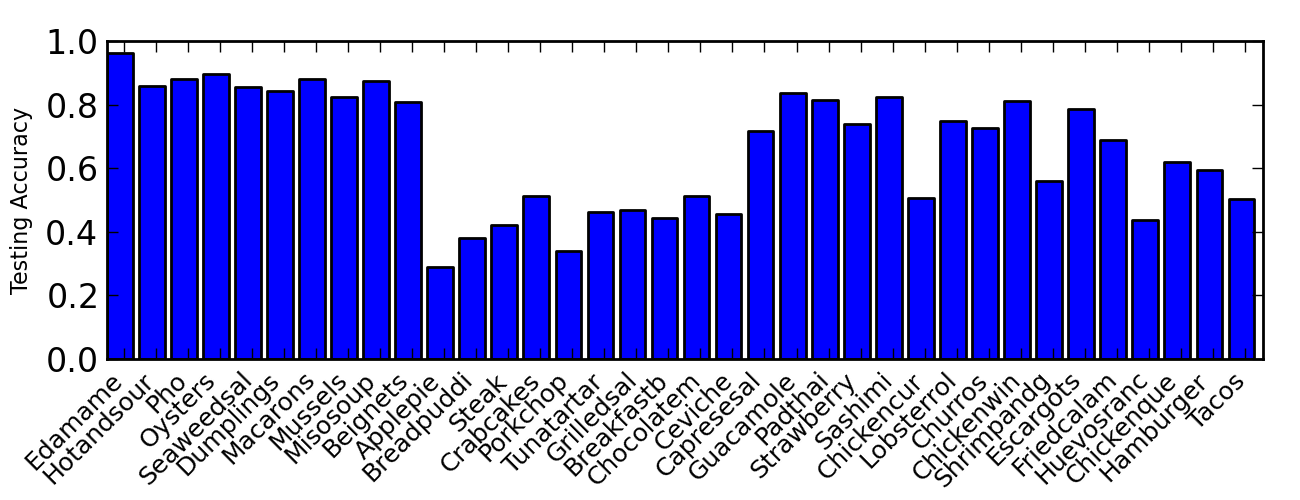}
\caption{Transfer learned Food-101 CNN accuracy across various classes
  in the dataset, presented for easy comparison with Figure 6 in
  Bossard et al. \shortcite{bossard14}. In general, this model finds
  the same classes difficult to classify as the models described in
  previous work, suggesting that some types of fine-grained
  distinctions are difficult for many models.}
\label{fig:food101}
\end{figure*}

To extend the scope of our investigation, we compile a dataset
consisting of images of food coupled with recipe titles from {\tt
  Yummly.com}. In this dataset, the title of a recipe is usually
several words long and can be thought of as a ``summary'' of the
image, rather than a direct description, as not all image content is
described in the caption. The image associated with ``garlic butter
shrimp,'' for instance, contains shrimp, a bowl, a lemon, and a human
hand, and the captioning algorithms must learn to pick out which items
are important to describe. Furthermore, there is less grammatical
structure present in this dataset.

We view this task as distinct from existing captioning tasks for three
reasons. First, the captions within Yummly are both short and
restricted; a caption in the Yummly setting has an average length of
4.5 words, which is very low compared to Flickr or MSCOCO settings
(both have an average of ~10 words per caption) and the vocabulary is
very small (see Figure \ref{fig:vocabsize}). Second, to address this
data fully, models must learn very fine-grained visual
distinctions. Compared to the broad ImageNet domain, the Yummly images
generally consist of some food item on a plate, coupled with several
words from a small vocabulary. Finally, this dataset contains a single
caption for each image, thus the learning task is more
difficult. Previous work \cite{hodosh2013framing} has emphasized the
importance of having multiple captions per image in a caption ranking
setting, though its unclear if similar observations extend to a
generation setting.

While we are only aware of the work of Malmaud et
al. \shortcite{malmaud2015s} that address food in a multimodal
fashion, Bossard et al. \shortcite{bossard14} compile the Food 101
dataset which generalizes and increases the scale of previous food
image datasets (i.e. Chen et al. \shortcite{chen2009pfid}, Yang et
al. \shortcite{yang2010food}). Their dataset includes 101k images of
101 types of foods and the task they address is classification.

\subsection{Choosing a CNN/RNN Architecture}

While substantial improvements have been made in terms of
classification accuracy on ImageNet using increasingly deep
architectures, we rely on the canonical neural network described in
Krizhevsky et al. \shortcite{alexnet} to generate our representations
in most of our experiments. The use of AlexNet in particular allows
for more direct comparison with previous work (i.e. Bossard et
al. \shortcite{bossard14}) and faster training time when compared to
other deep models. This is beneficial particularly because our
experiments are not specifically designed to produce state-of-the-art
results.

We perform 20 random parameter searches to determine decent parameter
settings using the Neuraltalk
\footnote{{\tt github.com/karpathy/neuraltalk}} library for all
captioning experiments, selecting parameter settings resulting in the
lowest validation set perplexity, unless specified otherwise. Settings
we take as fixed include a minimum vocabulary threshold of 5, weight
optimization using RMSprop \cite{tieleman2012lecture}, and a hidden
representation size of 256. We restrict our consideration to NIC
because we believe it to be representative of the state-of-the-art in
neural captioning. When we are evaluating models, we generate captions
using a beam search of width 20. For the recipe title prediction
evaluation, we include an end-of-caption token to avoid issues
relating to predicted zero length captions; this has the result of
artificially inflating evaluation metrics such that numerical
cross-dataset comparisons are not valid.

\subsection{Adapting the Food CNN through Transfer Learning}
To represent food images properly, we find it appropriate to learn a
model specific to the task of food recognition. Food-101
\cite{bossard14} consists of only 101K images, which is a relatively
low number of images to train a CNN from scratch. As such, we use a
set of ImageNet-trained weights as initializations for our training of
a CNN on the Food-101 classification task. This process is commonly
referred to as transfer learning
\cite{caruana1995learning,bengio2012deep}.

The intuition behind transfer learning in CNNs is that low-level
features learned early on in the base network (which are generally
observed to be color blob and Gabor features
\cite{yosinski2014transferable}) are useful to networks trained on
diverse classification tasks. Initializing the weights of the network
to weights successful in another classification task should allow
training of the new network to converge faster and to a better local
optimum than if random initializations were used.

In fact, for the Food-101 dataset, we achieve a rank-1 accuracy of
66.80\% when using transfer learning, when compared with the 56.40\%
rank-1 accuracy reported by Bossard et al. \shortcite{bossard14} using
the same AlexNet architecture; class-by-class accuracies are given in
Figure \ref{fig:food101} for comparison with previous work. Our network is
learned using only 100k iterations of the Caffe library at a reduced
learning rate, whereas training from scratch required Bossard et
al. 450k iterations. For our tuning process, we follow the guidelines
and parameter settings specified by the transfer learning example
distributed with Caffe.\footnote{{\tt https://github.com/BVLC/caffe}}

Once the network is tuned, we compute 4096 dimensional vector
representations for each image in Yummly dataset by extracting the
network activations in the final fully-connected layer.

\section{Yummly Dataset: Description and Baselines}
\begin{figure*}[ht]
\centering
\includegraphics[scale=1]{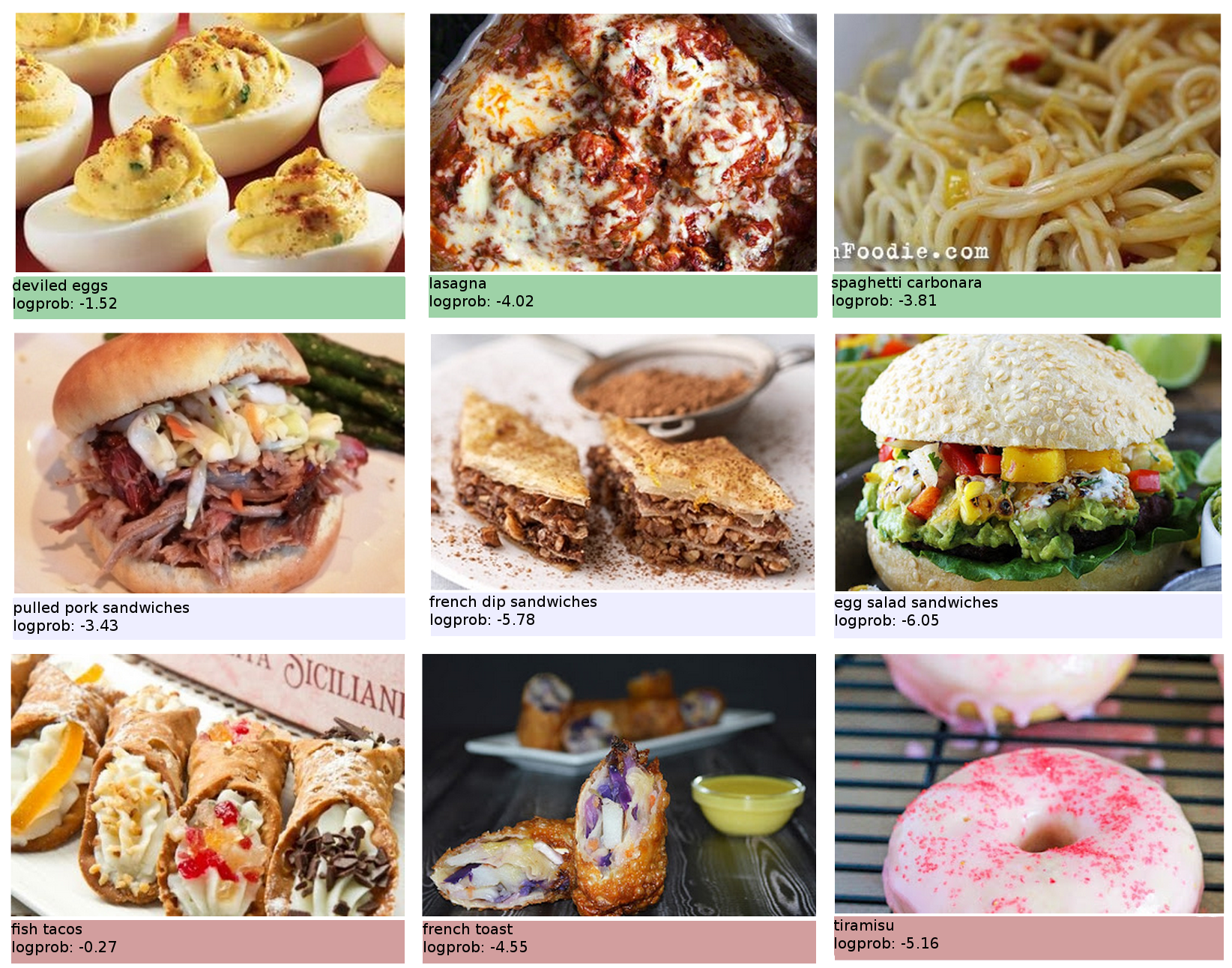}
\caption{Examples of the captioning system output on several
  images. The first row of images represents images that are well
  captioned. The second row represents different types of images the
  system believes to be sandwiches. The third row represents images
  that the system has captioned incorrectly.}
\label{fig:examples}
\end{figure*}

After establishing that a CNN could be transfer learned to classify
images of dishes at state-of-the-art performance, we were able to
shift our focus to caption generation in a food domain.

The food dataset we collect contains roughly 66K recipes, each
consisting of a single image-recipe pair. This data was taken from
{\tt Yummly.com}, a website that aggregates and performs analysis of
millions of recipes. Out of the 66K recipes, 6K are reserved for
testing, 6K are designated as a validation set, and the remaining 54K
are used for model training.

This dataset differs from the Flickr datasets and MSCOCO both in terms
of vocabulary and in terms of image content. The vocabulary size per
image is smaller than any of the other datasets by a wide margin (see
Figure \ref{fig:vocabsize}). While it's clear the vision task requires
more subtle distinction when compared to ImageNet, because the average
caption length is shorter, it's ambiguous as to whether or not the
Yummly language generation task is particularly ``fine-grained.''

\subsection{Baseline Results}

Table \ref{tab:baselines} presents some baseline results using the
algorithms listed. Common-3 predicts a reasonable ordering of the
three most common words (``with chicken and'') for all
captions. Nearest neighbor predicts the caption of nearest neighbor in
the transfer-learned 4096-dimensional embedding space. Common-Tri/Bi
predict the most common tri/bigram in our dataset (``macaroni and
cheese''/``ice cream'') for all images.

Across the board, and particularly for BLEU-\{2,3,4\} scores, the
caption generating programs outperform all baselines, which suggests
the proposed task is adequately framed. However, it is worth noting
that only roughly 300/6117 (roughly 5\%) of generated captions are
unique. This is rather low when compared with a representative result
for Flickr8k, a dataset of similar size, where 200/1000 (roughly 20\%)
of generated captions are unique. It might be possible to re-frame the
Yummly generation task as one of classification, however, it's not
obvious how one might drive a fixed set of labels. In a later section
we discuss whether or not only having one caption per image or other
dataset features is a contributing factor to this result.

\begin{table}
\centering
\begin{tabular}{|l||l|l|l|l|}
      \hline
       & B-1 & B-2 & B-3 & B-4\\\hline
      Com-3 & 14.2 & 2.7 & 0.8 & 0.0\\\hline
      N-Neigh & 20.5 & 2.5 & 0.6 & 0.0\\\hline
      Com-Tri & 30.4 & 6.5 & 3.4 & 2.2\\\hline
      Com-Bi & 35.4 & 8.9 & 5.2 & 0.0 \\\hline
      \pbox{20cm}{Karpathy and Li\\\shortcite{karpathy2014deep}} & 42.7 & 19.6 & 11.9 & 13.2\\\hline
      \pbox{20cm}{Vinyals et al.\\\shortcite{vinyals2014show}} & 46.2 & 23.1 & 14.8 & 10.2 \\\hline
\end{tabular}
    \caption{Yummly baseline BLEU-\{1,2,3,4\} scores for several baselines
      and two high performing language generation algorithms.}
    \label{tab:baselines}
\end{table}

\section{Image Representations}
\begin{figure*}[ht]
\centering
\begin{subfigure}{.5\textwidth}
  \centering
  \includegraphics[width=1\linewidth]{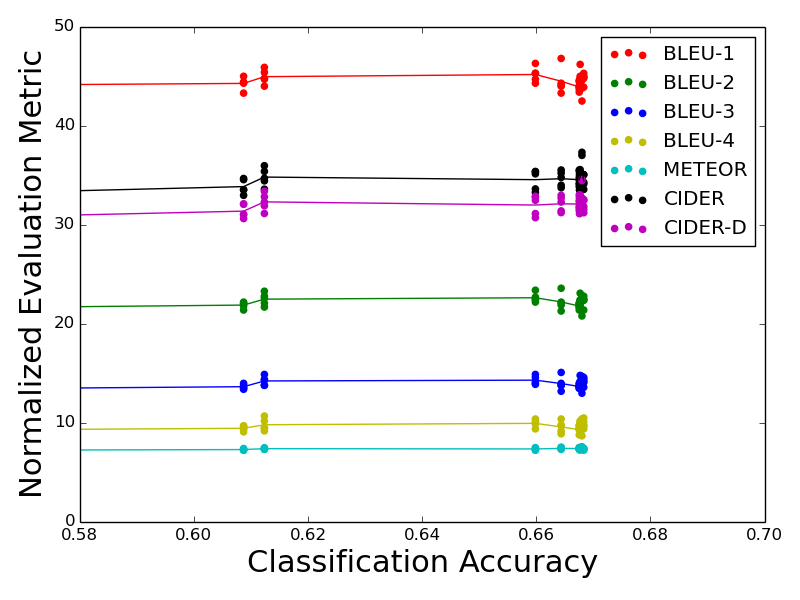}
  \caption{Yummly: Transfer learned domain}
\end{subfigure}%
\begin{subfigure}{.5\textwidth}
  \centering
  \includegraphics[width=1\linewidth]{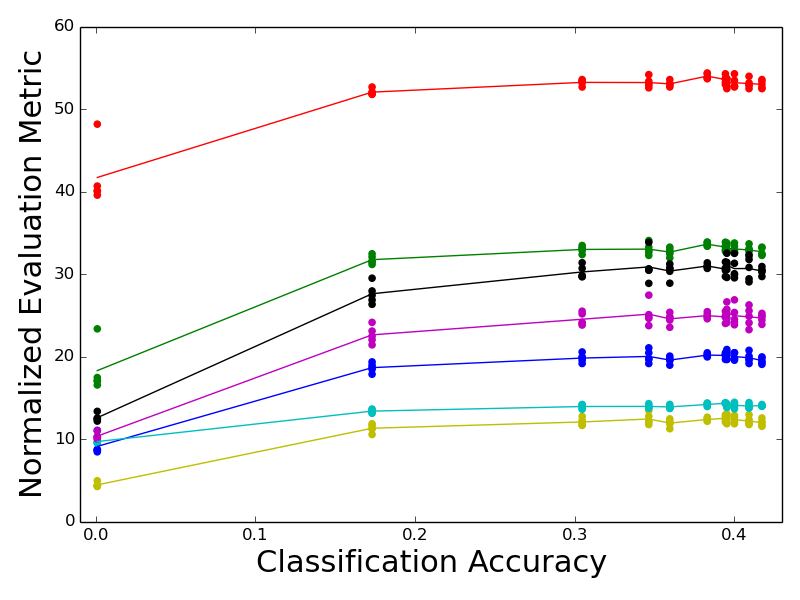}
  \caption{Flickr8k: Directly learned domain}
\end{subfigure}
\caption{Classification accuracy of CNN versus seven different
  normalized (100 is best possible) linguistic criteria for both the
  transfer learned (left) and directly learned (right) domains.}
\label{fig:results}
\end{figure*}

\subsection{Experiment Descriptions}
We vary image representation quality as follows: for the Flickr8k and
Flickr30k datasets, we compute the representations given by snapshots
of AlexNet taken mid-training on the ILSVRC2012 \cite{ILSVRC15}
task. We use snapshots taken at intervals of 10k from 0k (random
initialization) to 100k iterations. While this range of iterations is
before the model has entirely converged, the rank-1 classification
accuracy of the trained CNN over the ImageNet validation set increases
from roughly 0\% to over 40\% during this time (after the model
converges at 450k iterations, the rank-1 validation accuracy is
57\%). From the standpoint of examining representation quality, this
set of snapshots is important because this is likely where the network
is learning most of its layer-by-layer abstractions, and the behavior
of the network after 100k iterations can be extrapolated based on the
data we analyze here.

In a similar fashion, for Yummly we compute representations generated
by snapshots of the transfer learned network at intervals of 10k from
0k to 90k, though our starting point is a fully-converged CNN that
produces 57\% rank-1 accuracy on ImageNet's validation set.

We train 5 NIC models from a random initialization per CNN for
Flickr8k and Yummly, and 2-4 NIC models per CNN for Flickr30k. Every
data point described in the following section is the result of up to
six days of parallel computation using a modern 4/8-core machine. It
should be noted that test/validation accuracy of these CNNs is not
monotonically increasing with snapshot number. While the trend is that
training CNNs for more iterations results in higher accuracy, there is
some noise. For instance, for the Food-101 transfer learned CNN,
rank-1 test accuracy drops from 61\% to 60\% over the snapshots
extracted at 10k and 20k iterations respectively, before abruptly
jumping to 66\% testing accuracy in the next 10k iterations.
\subsection{Results}

We evaluate predicted captions using seven caption evaluation metrics,
namely, BLEU-\{1,2,3,4\}, METEOR, and CIDEr/CIDEr-D. Figure
\ref{fig:results} shows our main results for both the directly learned
and transfer learned domains. In both cases, all captioning metrics
appear to level off early, and do not improve significantly with
increased classification rate after a point. This suggests that weight
settings for a fixed CNN with higher classification rates are unlikely
to produce significantly better captions in terms of these seven
evaluation metrics, after a point.

To quantify this lack of improvement, for each dataset we select a CNN
that performs its associated visual classification task relatively
poorly, and compare it to all better-classifying CNNs. For Flickr8k,
for instance, we consider a CNN that produces 30.5\% rank-1 accuracy on
ImageNet's validation set, and compare its caption performance against
that of 8 ``better'' CNNs that achieve between 34.6\% and 41.7\%
accuracy; there are a total of 56 comparisons, in this case.

Though it is difficult to compute accurate statistics with only 5
observations in each group, we conduct three separate statistical
tests, each with different variance/normality
assumptions/efficiencies. The tests we perform are Students' t-test,
Mann-Whitney U-test, and Welch's unpaired t-test.

In the case of Flickr8k, there are very few significant differences
between the 30.5\%-CNN and more accurate CNNs. In fact, in 14/56 cases
(including half the time among BLEU-1/2 scores) the lower classifying
CNN actually produced better captions. The results significant at the
5\% level for any statistical test suggested that the 38\%-CNN
outperformed the 30.5\%-CNN in terms of BLEU-1/2, and that the
39.5\%-CNN outperformed the 30.5\%-CNN in terms of METEOR.

\begin{figure}
\centering
\includegraphics[width=.5\textwidth]{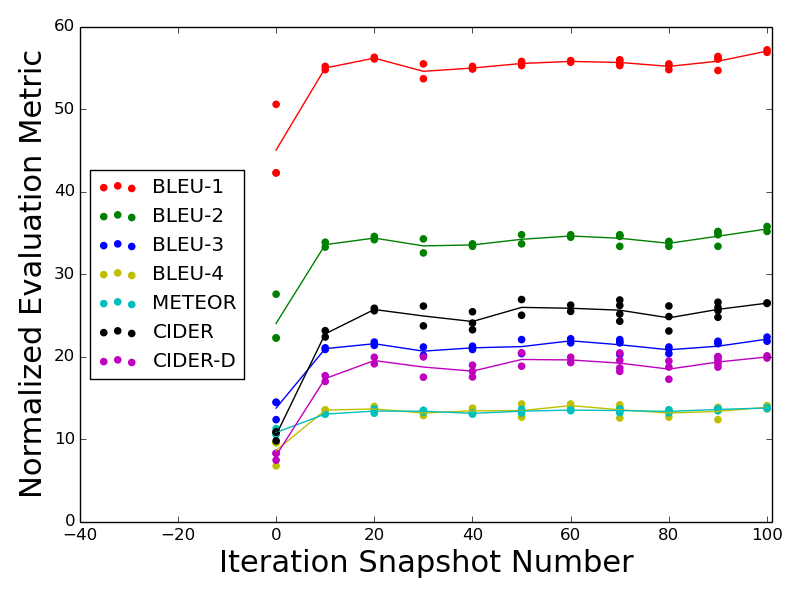}
\caption{Caption quality versus CNN iteration (in thousands of iters)
  that representations were derived from. It is clear that a caption
  quality saturation happens very early on, and there is little to no
  improvement in captions as the CNNs are trained for more time.}
\label{fig:f30k}
\end{figure}

The results for Flickr30k were very similar to the results for
Flickr8k. In Figure \ref{fig:f30k} we present results from this
dataset presented against CNN iteration number rather than CNN
classification accuracy. We modify the presentation of our data simply
to demonstrate that caption quality and iteration number (not just
testing/validation accuracy) are also apparently independent after a
point. No evidence of improvement was observed after the 30.5\%-CNN,
though only 2-4 observations per CNN could be made due to
computational restrictions.

In total, in the directly-learned domain (Flickr8k/30k) all metrics
appear to saturate after AlexNet reaches 30\% classification accuracy
over the ImageNet validation set. It is possible that training to
convergence could result in slightly higher quality captions. However,
our results indicate that efforts on ImageNet which result in less
than a roughly 10\% rank-1 classification accuracy increase for a
fixed network are likely not worth undertaking if one's end goal is
higher quality captions.

In the transfer learned domain, it is clear that domain adaptation
improves caption quality, even after a small number of iterations.
All statistical tests for all evaluation metrics indicate a highly
significant difference ($p < .01$) between captions generated by a CNN
trained directly on ImageNet, and one that has been transfer-learned
using Food-101 for just 10K iterations (producing a rank-1 testing
accuracy of 61.2\% on that dataset). After a point, however, we
observe the same independence of caption quality and classification
accuracy.

It seems that ``knowing more'' about the image does not help the RNN
generate more accurate captions after a point because the language
patterns it learns are sufficient. This result is akin to prior work
(e.g. Sutskever et al. \shortcite{sutskever2011generating}) which
demonstrates that RNNs are able to generate reasonable natural
language, given a relatively weak seeding signal. The ``weak'' signal
in this case is provided by image representations, rather than by a
short sequence of starting words.

\subsection{The Effect of Changing CNN Architectures}
Our analysis thus-far has focused on a single image model, AlexNet, for
extracting image representations. In this experiment, we compare the
captions generated on Flickr8k when using an improved CNN. We train 15
NIC models based on features extracted from a fully converged AlexNet,
and 15 NIC models based on features extracted form a fully converged
16-layer VGGNet \cite{simonyan2014very}. The former model produces a
57.1\% rank-1 accuracy over ImageNet's validation set, while the later
outperforms this mark, producing 75.6\% rank-1 validation accuracy. The
default train/valiation/test split of 6k/1k/1k images is used for
training.

Our results are summarized in Table \ref{tab:alexvgg}. In addition to
the seven caption evaluation metrics we've used in previous
experiments, this table also includes the proportion of the 1k
generated captions that are unique, and the train/validation
perplexities.

Counter-intuitively, we find that, despite producing 18\% lower rank-1
validation accuracy across ImageNet's validation set, AlexNet
generates \emph{better} captions than VGG net by all evaluation
metrics. Notably, the models using VGG features produce lower
perplexity across the validation split. Because we used validation
perplexity as a metric for hyperparameter selection, it's likely that
the VGG net models are overfitting to the particular Flickr8k
validation split we used. However, the AlexNet trained models do not
suffer a similar performance degradation. Here, it appears that not
overfitting to image features is more important than taking advantage
of very detailed image representations.

Our results from this experiment illustrate that better image
representations might actually cause models like NIC to become more
prone to overfitting. It's possible, too, that the early saturation of
caption quality observed in the previous sections could be primarily
due to overfitting. Future work would be well suited to evaluate
different methods of hyperparameter selection.

\subsection{One caption per image?}
We conclude with a final experiment to address one potential
shortcoming of domains similar to Yummly, where one is only able to
extract a single caption per image. Though Yummly differs from the
other datasets we explore in several ways (caption length/vocab size)
a fundamental question arises from its examination: for a fixed amount
of training data, is it better to have more captions per image, or
more images with single captions? In short, we hope to experimentally
examine Hodosh et al.'s \shortcite{hodosh2013framing} suggestion that
having multiple captions per image is vital.

To address this question, we use Flickr30k, which provides five
captions per image. We subset this dataset in two ways. In the first,
we remove 4 captions randomly from each image in the training set, but
keep all images (the ``more images'' method). In the second, we
randomly remove 80\% of training images, but keep all 5 captions for
the remaining (the ``more captions'' method). This subsetting scheme
is such that the overall number of image/caption pairs is the same
between both methods, but the training data is of a different form.

We extract image representations from the ImageNet CNN at 100k
iterations (which produces roughly 40\% rank-1 classification accuracy
over the ImageNet validation set) and train NIC on 6 random datasets
constructed via the ``more images'' subsetting method, and 7 random
datasets constructed via the ``more captions'' subsetting
method. Finally, we generate captions and compare performance.  A good
hyperparameter setting for Flickr30k is borrowed from the random
search conducted over the whole dataset experiments described in the
previous section.

Our findings, summarized in Table \ref{tab:morecap}, generally align
with the accepted notion that having more captions and less images is
better than having more images with single captions. For all seven
evaluation metrics, the mean score for the models trained on the
``more captions'' datasets was greater than the mean score for the
models trained on the ``more images'' datasets, and the results were
significant at the 5\% level for all three statistical tests in the
case of BLEU-1 and BLEU-2. Interestingly, for CIDEr/CIDEr-D, the
results were somewhat significant (all 6 p-values less than $.15$) but
the results for METEOR were the least significant (all 3 p-values
greater than $.94$).

The validation perplexity of the ``more images'' method is lower when
compared to the more captions method, whereas the training perplexity
is higher. Despite the fact that the output captions are better
overall, this is an indication that having multiple captions per image
can actually make NIC more prone to overfitting.

Finally, the NIC models trained on the ``more caption'' subsets
produced higher proportions of unique captions on the test set. This
suggests that the single-caption per image feature of the Yummly
dataset contributed to a lack of caption innovation.

Despite only having one caption per image, however, NIC was still able
to produce good results on the single-captioned subsets. This
indicates that quality captioning datasets can be built with only one
caption per image. The number of additional images one needs to gather
to compensate for this feature, however, is likely greater than the
number of captions one would need to add to existing images.

\begin{table}[t]
\centering
\begin{tabular}{|l|l|l|}
  \hline
     & AlexNet & VGG\\\hline
\pbox{20cm}{Top-1 ImageNet\\Val Acc} &
  57.1\% & 75.6\%\\\hline\hline B-1 & 54.187 & 53.913\\\hline B-2 & 33.967 &
  33.527\\\hline B-3** & 20.640 & 20.007\\\hline B-4** & 12.833 &
  12.213\\\hline METEOR & 14.559 & 14.559\\\hline CIDEr & 32.416 &
  31.362\\\hline CIDEr-D* & 26.200 & 25.242\\\hline\hline
  \pbox{20cm}{Proportion\\Unique***} & 20.5\% & 17.0\%\\\hline
  \pbox{20cm}{Training\\Perplexity***} & 10.79 & 11.04\\\hline
  \pbox{20cm}{Validation\\Perplexity***} & 17.84 & 17.66\\\hline
\end{tabular}
\caption{Effect on caption quality when using the fully converged
  AlexNet and VGGNet on Flickr8k. Significance for all 3 statistical
  tests that there was a true difference between the subsetting
  techniques: ***$p<.001$, **$p<.01$, *$p<.05$}
    \label{tab:alexvgg}
\end{table}

\begin{table}[t]
\centering
\begin{tabular}{|l|l|l|}
      \hline
       & More Captions & More Images\\\hline
      B-1** & 55.167 & 54.243\\\hline
      B-2* & 33.567 & 32.814\\\hline
      B-3 & 20.633 & 20.300\\\hline
      B-4 & 13.133 & 13.014\\\hline
      METEOR & 13.105 & 13.096\\\hline
      CIDEr & 21.428 & 20.418\\\hline
      CIDEr-D & 16.350 & 15.550\\\hline\hline
      \pbox{20cm}{Proportion\\Unique**} & 14.8\% & 9.96\%\\\hline
      \pbox{20cm}{Training\\Perplexity**} & 14.69 & 16.01\\\hline
      \pbox{20cm}{Validation\\Perplexity*} & 25.86 & 25.33\\\hline
\end{tabular}
    \caption{Evaluations for the NIC models trained on subsets of
      Flickr30k containing more captions (5 captions per image, 1/5
      the total number of images) and more images (1 caption per
      image, all training images). Significance for all 3 statistical
      tests that there was a true difference between the subsetting
      techniques: **$p<.01$, *$p<.05$}
    \label{tab:morecap}
\end{table}

\section{Conclusion}
We demonstrate the relationship between CNN classification accuracy
and the quality of captions generated by a state of the art neural
captioning algorithm. Training increasingly accurate image classifiers
does not lead to better captions, after a point. This early saturation
of caption quality is an indication that the performance of neural
caption generating algorithms likely cannot be increased directly by
producing more accurate CNNs. Furthermore, many of the apparently
highly-specific generated captions output by models like NIC are
likely due to language models capturing coarse grained information
and generating corresponding plausible natural language sequences.

The role of overfitting to image features is difficult to quantify. On
one hand, there is extra information contained in image
representations that NIC, for instance, does not take advantage of,
and even commonly overfits to. However, it's not clear that this
extra, fine-grained information is even worth taking into account. The
success of models that generate language based on discretized image
representations (e.g. \cite{young2014image}) demonstrates that
algorithms are capable of state-of-the-art performance without
consideration of rich, real-valued vector features. It's likely that
these types of models are less prone to overfitting, as well.
\section{Acknowledgments}

We would like to thank Jason Yosinski for providing his AlexNet
training snapshots/insights and Gregory Druck for his help with
compiling the data collected from \texttt{Yummly.com}. We would also
like to thank Serge Belongie, Lillian Lee, Abby Lewis, David Mimno,
Xanda Schofield, the anonymous reviewers, and the students in the
Spring 2015 iteration of CS6670 for their helpful discussions and
comments.

\bibliographystyle{acl}
{\footnotesize
\bibliography{ref}}

\end{document}